%% file: main.tex
\pdfoutput=1
\documentclass[10pt,twocolumn,letterpaper]{article}

\usepackage{cvpr}              

\usepackage{graphicx}
\usepackage{amsmath}
\usepackage{amssymb}
\usepackage{booktabs}
\usepackage{eucal}
\usepackage{adjustbox}
\usepackage[accsupp]{axessibility}  

\usepackage[pagebackref,breaklinks,colorlinks]{hyperref}
\usepackage{multirow}
\usepackage{colortbl}
\usepackage[dvipsnames]{xcolor}
\usepackage{tabularx}
\usepackage{arydshln}
\usepackage{hhline}
\usepackage{amssymb}
\newcommand{\xmark}{\text{\sffamily X}}
\definecolor{LightCyan}{rgb}{0.88,1,1}

\usepackage[capitalize]{cleveref}
\usepackage{makecell}
\crefname{section}{Sec.}{Secs.}
\Crefname{section}{Section}{Sections}
\Crefname{table}{Table}{Tables}
\crefname{table}{Tab.}{Tabs.}
\definecolor{lightgreen}{HTML}{D8ECD1}

\def\cvprPaperID{5460} 
\def\confName{CVPR}
\def\confYear{2023}

\begin{document}
\title{Neural Map Prior for Autonomous Driving}
\author{Xuan Xiong$\phantom{}^{1}$\hspace{15pt}
Yicheng Liu$\phantom{}^{1}$\hspace{15pt}
Tianyuan Yuan$\phantom{}^{2}$\hspace{15pt}
Yue Wang$\phantom{}^{3}$\hspace{15pt}
Yilun Wang$\phantom{}^{2}$\hspace{15pt}
Hang Zhao$\phantom{}^{2,1}$\thanks{Corresponding at: \texttt{hangzhao@mail.tsinghua.edu.cn}} \vspace{8pt}\\
$\phantom{}^1$Shanghai Qi Zhi Institute\hspace{5pt}
$\phantom{}^2$IIIS, Tsinghua University\hspace{5pt}
$\phantom{}^3$MIT\hspace{5pt}
}

\maketitle
\input{abstract.tex}
\input{introduction.tex}
\input{figs/fig2.tex}
\input{related_works.tex}
\input{method}
\input{experiment}
\input{conclusion.tex}

\subsubsection*{Acknowledgments}
This work is in part supported by Li Auto. 
We thank Binghong Yu for proofreading the final manuscript.

{\small
\bibliography{main.bib}
\bibliographystyle{ieee_fullname}
}
\end{document}

%% file: abstract.tex
\begin{abstract}
High-definition (HD) semantic maps are crucial in enabling autonomous vehicles to navigate urban environments. The traditional method of creating offline HD maps involves labor-intensive manual annotation processes, which are not only costly but also insufficient for timely updates. Recent studies have proposed an alternative approach that generates local maps using online sensor observations. However, this approach is limited by the sensor's perception range and its susceptibility to occlusions.
In this study, we propose \textbf{Neural Map Prior (NMP)}, a neural representation of global maps. This representation automatically updates itself and improves the performance of local map inference. Specifically, we utilize two approaches to achieve this. Firstly, to integrate a strong map prior into local map inference, we apply cross-attention, a mechanism that dynamically identifies correlations between current and prior features. Secondly, to update the global neural map prior, we utilize a learning-based fusion module that guides the network in fusing features from previous traversals. Our experimental results, based on the nuScenes dataset, demonstrate that our framework is highly compatible with various map segmentation and detection architectures. It significantly improves map prediction performance, even in challenging weather conditions and situations with a longer perception range. To the best of our knowledge, this is the first learning-based system for creating a global map prior.
\end{abstract}

%% file: introduction.tex
\section{Introduction}
\label{sec:intro}
Autonomous vehicles require high-definition (HD) semantic maps to accurately predict the future trajectories of other agents and to navigate urban environments safely. However, the majority of these vehicles rely on costly and labor-intensive pre-annotated offline HD maps. These maps are constructed through a complex pipeline involving multiple LiDAR scanning trips with survey vehicles, global point cloud alignment, and manual annotation of map elements. Despite the high precision of these offline mapping solutions, their scalability is constrained, and they do not support timely updates in response to changing road conditions. As a result, autonomous vehicles may operate based on outdated maps, which could compromise driving safety.

\begin{figure}[t]
        \hspace{-0.06cm}
         \includegraphics[height=1.05\linewidth]{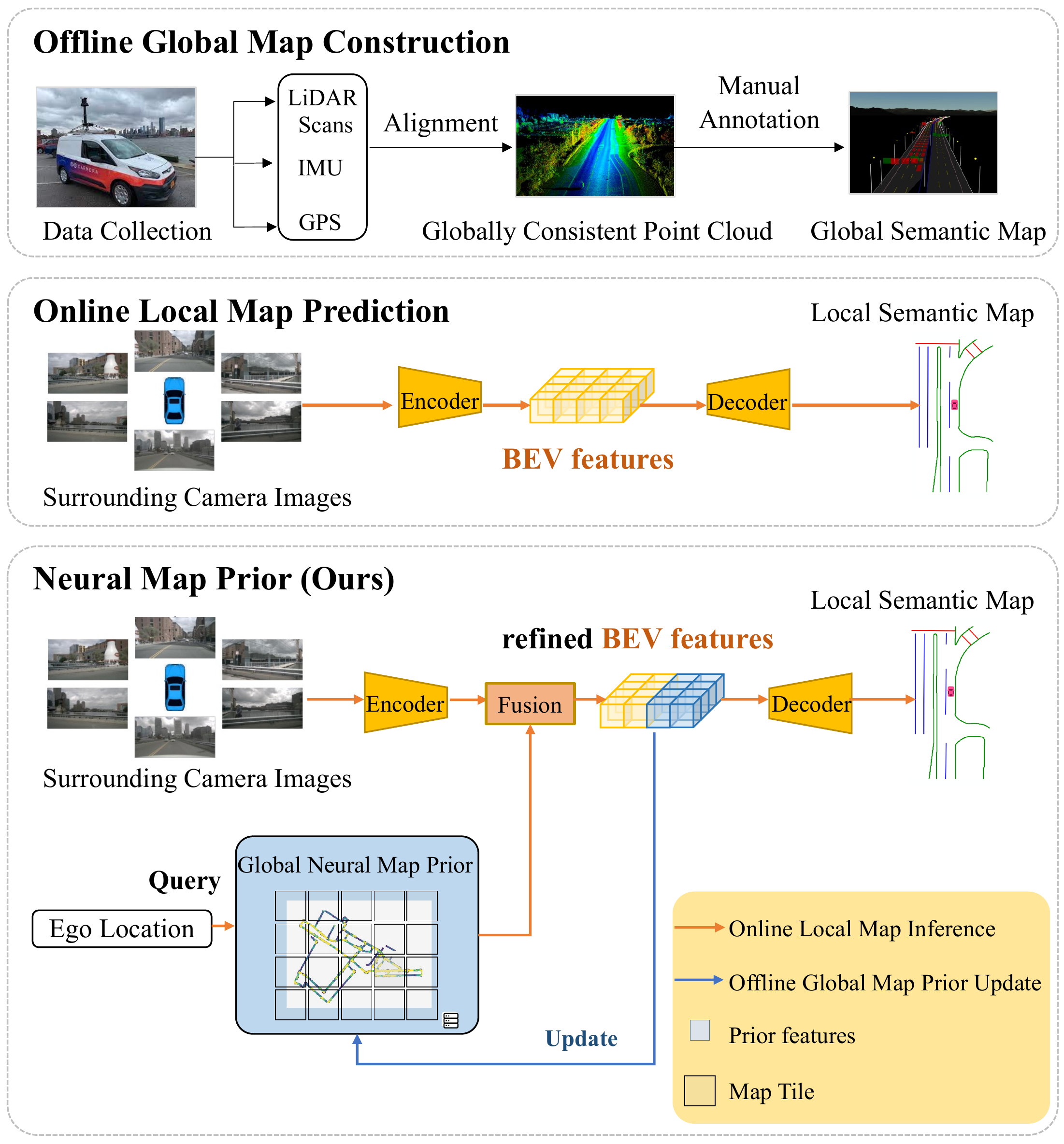}
           \caption{\textbf{Comparison of semantic map construction methods.} Traditional offline semantic mapping pipelines (the first row) involve a complex manual annotation pipeline and do not support timely map updates. Online HD semantic map learning methods (the second row) rely entirely on onboard sensor observations and are susceptible to occlusions. We propose the Neural Map Prior (NMP, the third row), an innovative neural representation of global maps designed to aid onboard map prediction. NMP is incrementally updated as it continuously integrates new observations from a fleet of autonomous vehicles.}
           \label{fig:fig1}
\vspace{-1em}
\end{figure}

Recent research has explored alternative methods for constructing HD semantic maps using onboard sensor observations, such as camera images and LiDAR point clouds~\cite{li2021hdmapnet,liu2022vectormapnet,liao2022maptr}. These methods typically use deep learning techniques to infer map elements in real-time, thus addressing the issue of map updates associated with offline maps. Nevertheless, the quality of these inferred maps is generally inferior when compared to pre-constructed global maps. And this quality can degrade further under unfavorable weather conditions or in occluded scenarios. The comparison of different semantic map construction methods is provided in Figure ~\ref{fig:fig1}.

In this study, we present Neural Map Prior (NMP), a novel hybrid mapping solution that combines the best of both worlds. NMP leverages neural representations to build and update a global map prior, thereby enhancing local map inference performance. The NMP methodology consists of two important stages: \textbf{global map prior update} and \textbf{local map inference}. The global map prior is automatically developed by aggregating sensor data from a fleet of self-driving cars. Onboard sensor data and the global map prior are then integrated into the local map inference process, which subsequently refines the map prior. These procedures are interconnected in a feedback loop that grows stronger as more data are collected from vehicles traversing the roads daily. One example is shown in Figure ~\ref{fig:teaser}.

Technically, the global NMP is defined as sparse map tiles, where each tile corresponds to a specific real-world location and starts in an empty state. For each online observation from an autonomous vehicle, a neural network encoder first extracts local bird's-eye view (BEV) features. These features are then refined using the corresponding NMP prior features, derived from the global NMP's map tile. The refined BEV feature enables us to better infer the local semantic map and update the global NMP. As the autonomous vehicles traverse through various scenes, the local map inference phase and the global map prior update step mutually reinforce each other. This iterative process results in improved quality of the predicted local semantic map and maintains a more complete and up-to-date global NMP.

We demonstrate that our NMP can be easily applied to various state-of-the-art HD semantic map learning methods, effectively enhancing their accuracy. Through experiments conducted on the nuScenes dataset, our pipeline showcases remarkable performance improvements, including a +\textbf{4.32} mIoU for HDMapNet, +\textbf{5.02} mIoU for LSS, +\textbf{5.50} mIoU for BEVFormer, and +\textbf{3.90} mAP for VectorMapNet.

To summarize, our contributions are as follows:
\begin{enumerate}
    \item We propose a novel mapping paradigm, Neural Map Prior, which integrates the maintenance of offline global maps and the inference of online local maps. Notably, the computational and memory resources required by our approach's local map inference are comparable to previous methods.
    \item We propose current-to-prior attention and Gated Recurrent Unit modules. These are adaptable to mainstream HD semantic map learning methods and effectively enhance their map prediction performance.
    \item We conduct a comprehensive evaluation of our method on the nuScenes dataset, considering different map elements and four map segmentation/detection architectures. The results demonstrate consistent and significant improvements. Moreover, our approach demonstrates substantial progress in challenging scenarios, such as bad weather conditions and longer perception ranges.
\end{enumerate}

%% file: figs/fig2.tex
\label{sec:review}
\begin{figure*}[htb]
        \centering
         \includegraphics[height=6cm]{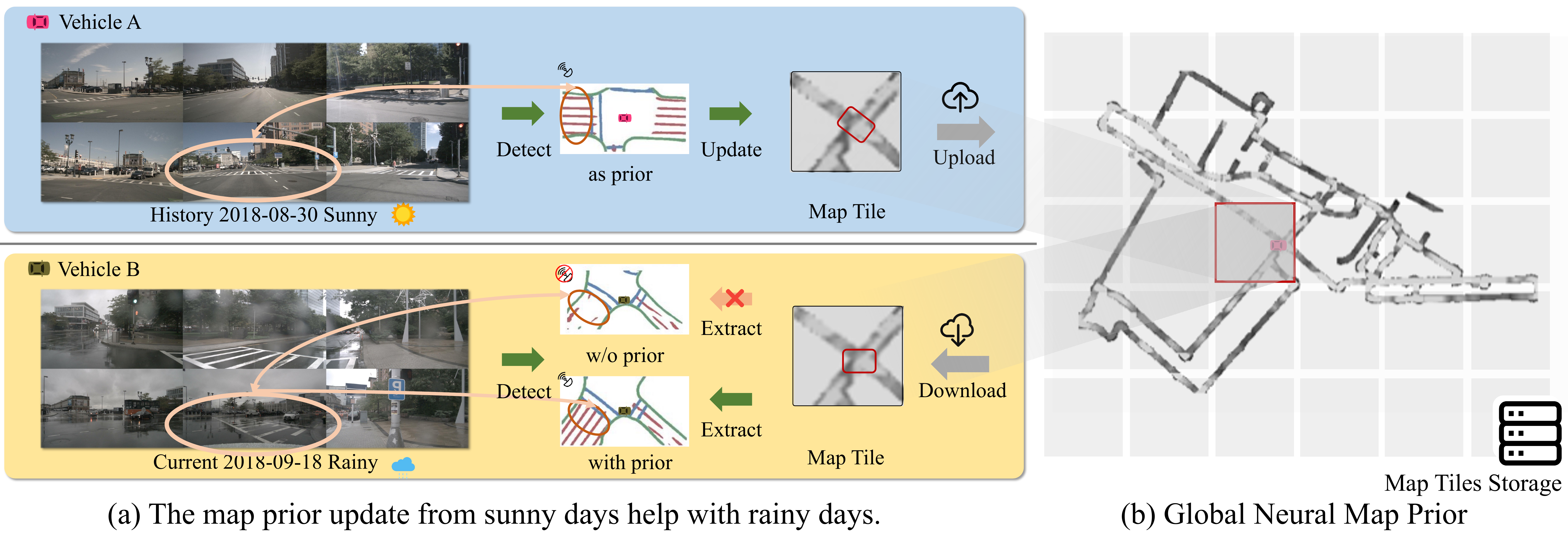}
          \caption{\textbf{Demonstration of NMP for autonomous driving in adverse weather conditions.} Ground reflections during rainy days make online HD map predictions harder, posing safety issues for an autonomous driving system. NMP helps to make better predictions, as it incorporates prior information from other vehicles that have passed through the same area on sunny days. 
          }
          \label{fig:teaser}
\end{figure*}

%% file: related_works.tex
\section{Related Works}\label{sec:related}
\noindent\textbf{LiDAR SLAM-Based Mapping.} Autonomous driving system requires an understanding of road map elements, including lanes, pedestrian crossing, and traffic signs, to navigate the world. Such map elements are typically provided by pre-annotated High-Definition (HD) semantic maps in existing pipelines~\cite{rong2020lgsvl}. Most current HD semantic maps are manually or semi-automatically annotated on LiDAR point clouds of the environment, merged from LiDAR scans collected from survey vehicles equipped with high-end GPS and IMU. SLAM algorithms are the most commonly used algorithms to fuse LiDAR scans into a highly accurate and consistent point cloud. First, to match LiDAR data at two nearby timestamps, pairwise alignment algorithms such as ICP~\cite{besl1992method}, NDT~\cite{biber2003normal}, and their variants~\cite{segal2009Generalized} are employed, using either semantic~\cite{yu2015semantic} or geometry information~\cite{pomerleau2015review}. Second, accurately estimating the poses of the ego vehicles is critical for building a globally consistent map and is formulated as either a non-linear least-square problem~\cite{lawson1995solving} or a factor graph~\cite{dellaert2012factor}. Yang et al.~\cite{yang2018robust} presented a method for reconstructing city-scale maps based on pose graph optimization under the constraint of pairwise alignment factors. To reduce the cost of manual annotation of semantic maps, Jian et al. ~\cite{jiao2018machine} proposed several machine-learning techniques for extracting static elements from fused LiDAR point clouds and cameras. However, maintaining an HD semantic map remains a laborious and costly process due to the requirement for high precision and timely updates. In this paper, we propose using neural map priors as a novel mapping paradigm to replace human-curated HD maps, supporting timely updates to the global map prior and enhancing local map learning, potentially making it a more scalable solution for autonomous driving.\\
\noindent\textbf{Semantic Map Learning.}
Semantic map learning constitutes a fundamental challenge in real-world map construction and has been formulated as a semantic segmentation problem in ~\cite{mattyus2015enhancing}. Various approaches have been employed to address this issue, including aerial images in ~\cite{mattyus2016hd}, LiDAR point clouds in ~\cite{yang2018hdnet}, and HD panoramas in ~\cite{wang2016torontocity}. To enhance fine-grained segmentation performance, crowdsourcing tags have been proposed in ~\cite{wang2015holistic}. Recent studies have concentrated on deciphering BEV semantics from onboard camera images ~\cite{lu2019monocular,yang2021projecting} and videos ~\cite{can2020understanding}. Relying solely on onboard sensors for model input poses a challenge, as the inputs and target map belong to different coordinate systems. Cross-view learning methodologies, such as those found in ~\cite{philion2020lift,pan2020cross,li2021hdmapnet,zhou2022cross,wang2022detr3d,chen2022futr3d,saha2022translating,roddick2020predicting}, exploit scene geometric structures to bridge the gap between sensor inputs and BEV representations. Our proposed method capitalizes on the inherent spatial properties of BEV features as a neural map prior, making it compatible with a majority of BEV semantic map learning techniques. Consequently, this approach holds the potential to enhance online map prediction capabilities.\\
\noindent\textbf{Neural Representations.}
Recently, advances have been made in neural representations~\cite{meschederOccupancyNetworksLearning2018,parkDeepSDFLearningContinuous2019,saitoPIFuPixelAlignedImplicit2019,jiangLocalImplicitGrid2020,yarivMultiviewNeuralSurface2020,liu2020neural}. NeuralRecon~\cite{sun2021neuralrecon} presents an approach for implicit neural 3D reconstruction that integrates reconstruction and fusion processes. Unlike traditional methods that first estimate depths and subsequently perform fusion offline. Similarly, our work learns neural representation by employing the encoded image features to predict the map prior through a neural network.

%% file: method.tex
\section{Neural Map Prior}
\label{sec:method}
The aim of this work is to improve local map estimation performance by leveraging a global neural map prior. To achieve this, we propose a pipeline, depicted in Figure~\ref{fig:model_architecture}, which is specifically designed to concurrently train both the global map prior update and local map learning with integrating a fusion component. Moreover, we address the memory-intensive challenge associated with storing features of urban streets by introducing a sparse tile format for the global neural map prior, as detailed in Section~\ref{subsec:exp_tile}.\\

\noindent\textbf{Problem Setup.}
Our model operates on typical autonomous driving systems equipped with an array of onboard sensors, such as surround-view cameras and GPS/IMU, for precise localization. We assume a single-frame setting, similar to \cite{li2021hdmapnet}, which adopts a BEV encoder-decoder model for inferring local semantic maps. The BEV encoder is denoted as $F_{\mathrm{enc}}$, and the decoder is denoted as $F_{\mathrm{dec}}$. Additionally, we create and maintain a global neural map prior $p^{g}\in\mathbb{R}^{H_G \times W_G \times C}$, where $H_G$ and $W_G$ represent the height and width of the city, respectively. Each observation consists of input from the surrounding cameras $\mathbf{I}$ and the ego vehicle's position in the global coordinate system $\mathbf{G}_{\mathrm{ego}}\in\mathbb{R}^{4 \times 4}$. We can transform the local coordinate of each pixel of the BEV, denoted as $l_{\mathrm{ego}} \in\mathbb{R}^{H \times W\times 2}$ (where $H$ and $W$ denote the size of the BEV features), to a fixed global coordinate system using $\mathbf{G}_{\mathrm{ego}}$. This transformation results in $p_{\mathrm{ego}}\in\mathbb{R}^{H \times W\times 2}$. Initially, we acquire the online BEV features $o=F_{\mathrm{enc}}(\mathbf{I})\in\mathbb{R}^{H \times W \times C}$, where $C$ represents the network's hidden embedding size. We then query the global prior $p^g$ using the ego position $p_{\mathrm{ego}}$ to obtain the local prior BEV features $p^l_{t-1}\in\mathbb{R}^{H \times W\times C}$. A fusion function is subsequently applied to the online BEV features and the local prior BEV features to yield refined BEV features:
\begin{equation}
f_{\text{refine}} = F_{\text{fuse}} (o, p^l_{t-1}), f_{\text{refine}}\in\mathbb{R}^{H \times W \times C}. 
\end{equation}
Finally, the refined BEV features are decoded into the final map outputs by the decoder $F_{\mathrm{dec}}$. Simultaneously, the global map prior $p^g$ is updated using $f_{\text{refine}}$. The global neural network prior acts as an external memory, capable of incrementally integrating new information and simultaneously offering knowledge output. This dual functionality ultimately leads to improved local map estimation performance.

\input{figs/fig3.tex}
\subsection{Local Map Learning}
\label{subsec:localinference}
In order to accommodate the dynamic nature of road networks in the real world, advanced online map learning algorithms have recently been developed. These methods generate semantic map predictions based solely on data collected by onboard sensors. In contrast to earlier approaches, our proposed method incorporates neural priors to bolster accuracy. As road structures on maps are subject to change, it is imperative that recent observations take precedence over older ones. To emphasize the importance of current features, we introduce an asymmetric fusion strategy that combines current-to-prior attention and gated recurrent units.\\

\noindent\textbf{Current-to-Prior (C2P) Cross Attention.}
We introduce the current-to-prior cross-attention mechanism, which employs a standard cross-attention approach~\cite{liu2021swin} to operate between current and prior BEV features. Concretely, We divide each BEV feature into patches and add them with a set of learnable positional embeddings, which will be described subsequently. Current features produce queries, while prior features produce keys and values. A standard cross-attention is then applied, succeeded by a fully connected layer. Ultimately, we assemble the output queries to derive the refined BEV features, which maintain the same dimensions as the input current features. The resulting refined BEV features are expected to exhibit superior quality compared to both prior and current features.\\

\noindent\textbf{Positional Embedding.}
It has been observed that the accuracy of predicted maps declines as the distance from the ego vehicle increases. To address this issue, we propose the integration of position embeddings, a set of grid-shaped learnable parameters, into the fusion module. The aim is to augment the spatial awareness of the fusion module regarding the feature positions, empowering it to learn to trust the current features closer to the ego vehicle and rely more on the prior features for distant locations. Specifically, two position embeddings are introduced: $PE_{p} \in \mathbb{R}^{H \times W \times C}$ for the prior features and $PE_{c} \in  \mathbb{R}^{H \times W \times C}$ for the current features, respectively, before the fusion module $F_{\mathrm{fuse}}$. Here, $H$ and $W$ represent the height and width of the BEV features. These embeddings provide spatial awareness to the fusion module, effectively allowing it to assimilate information from varying feature distances and locations.

\subsection{Global Map Prior Update}
\label{subsec:GRU}
To update the global map prior with the refined features generated by the C2P attention module, an auxiliary module is introduced, devised to attain a balanced ratio between the current and prior features. This process is illustrated in Figure~\ref{fig:model_architecture}. Intuitively, the module regulates the updating rate of the global map prior. A high update rate may lead to corruption of the global map prior due to suboptimal local observations, while a low update rate may result in the global map prior's inability to promptly capture changes in road conditions. Therefore, we introduce a 2D convolutional variant of the Gated Recurrent Unit~\cite{chung2014empirical} module into NMP, serving to balance the updating and forgetting ratio. Local map prior features $p_{t-1}^{l}$, updated at $t-1$, are extracted from the global neural map prior $p_{t-1}^g$. The refined features generated by the C2P attention module are denoted as $o^{'}$. Integrating $o^{'}$ with the local prior features $p_{t-1}^l$, the GRU yields the new prior features $p_t^l$ at time $t$. Subsequently, these features are passed through the decoder to predict the local semantic map and the global neural map prior $p_{t}^g$ is updated at the corresponding location by directly replacing them with $p_{t}^l$. Let $z_t$ denote the update gate, $r_t$ the reset gate, $\sigma$ the sigmoid function, $w_{*}$ the weight for 2D convolution, and $\odot$ the Hadamard product. Via the following operations, the GRU fuses $o^{'}$ with the prior feature $p_{t-1}^l$: 
\begin{align}
    \begin{split}
    &z_t = \sigma(\text{Conv2D}([p^l_{t-1}, o^{'}], w_z)) \\
    &r_t = \sigma(\text{Conv2D}([p^l_{t-1}, o^{'}], w_r)) \\
    &\tilde{p^l_{t}} = \tanh(\text{Conv2D}([r_t \odot p^l_{t-1}, o^{'}], w_h)) \\
    & p^l_{t} = (1 - z_t) \odot p^l_{t-1} + z_t \odot \tilde{p^l_{t}}
    \end{split}
\end{align}
Within the GRU, the update gate $z_t$ and reset gate $r_t$ are instrumental in determining the fusion of information from the previous traversal (\ie, prior feature $p^l_{t-1}$) with the current BEV feature $o^{'}$. Furthermore, they govern the incorporation of information from the current BEV feature $o^{'}$ into the global map prior feature $p^l_{t}$. GRU enables the model to better adapt to various road conditions and mapping scenarios more effectively.

%% file: figs/fig3.tex
\begin{figure*}[h]
\vspace{-0.0cm}
        \centering
         \includegraphics[height=8cm]{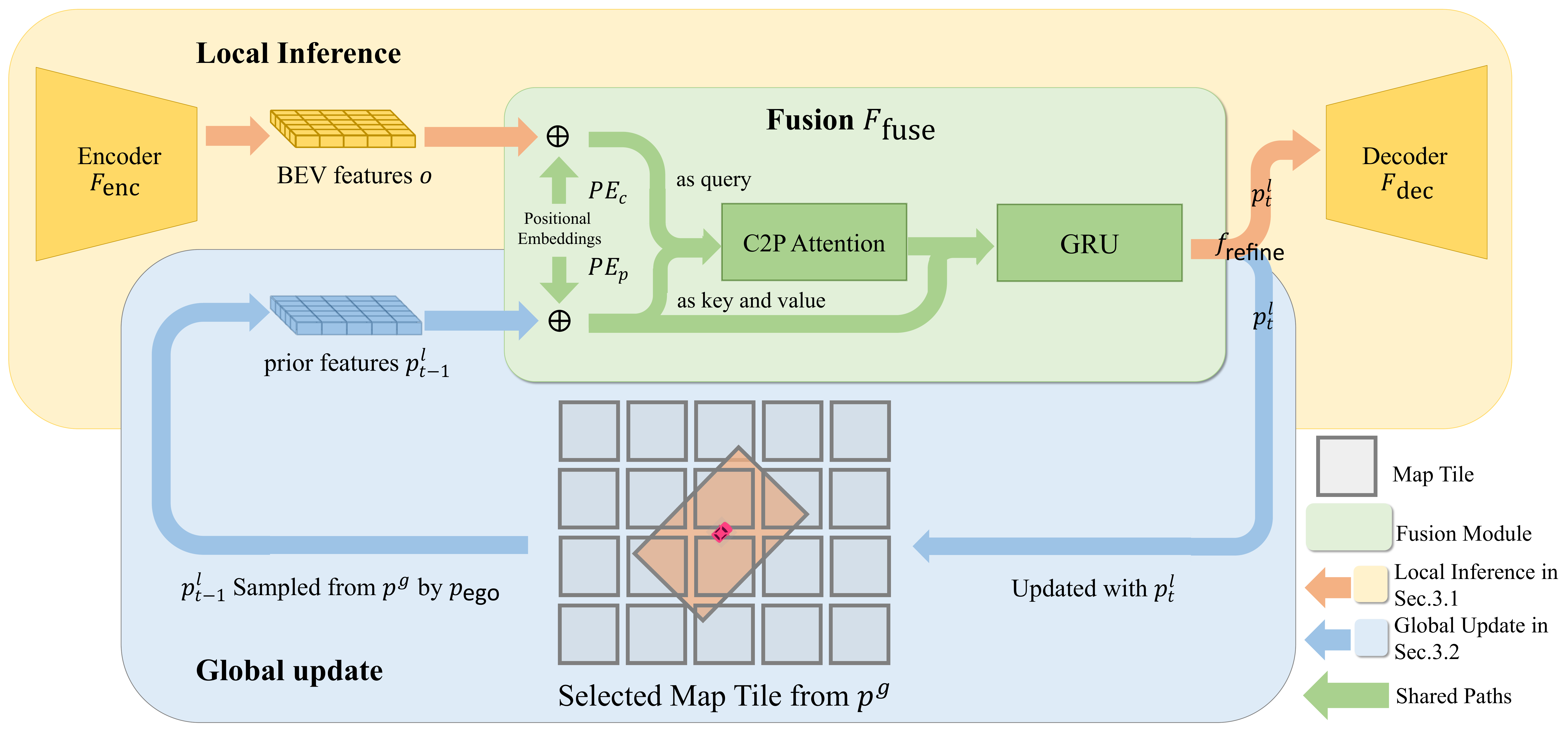}
          \caption{\textbf{The model architecture of NMP.}  The top yellow box illustrates the online HD map learning process, which takes images as input and processes them through a BEV encoder and decoder to generate map segmentation results. Within the green box, customized fusion modules—comprising C2P attention and GRU—are designed to effectively integrate prior map features between the encoder and decoder, subsequently decoded to produce the final map predictions. In the bottom blue box, the model queries map tiles that overlap with the current BEV feature from storage. After the update, the neural map is returned to the previously extracted map tiles.}
          \label{fig:model_architecture}
\end{figure*}

%% file: experiment.tex
\section{Experiments}
\label{sec:expr}

\paragraph{Datasets.} 
We validate our NMP on the nuScenes dataset~\cite{caesar2020nuscenes}, a large-scale autonomous driving benchmark that includes multiple traversals with precise localization and annotated HD map semantic labels. The NuScenes dataset contains 700 scenes in the \textit{train}, 150 in the \textit{val}, and 150 in the \textit{test}. Data were collected using a 32-beam LiDAR operating at 20 Hz and six cameras offering a 360-degree field of view at 12 Hz. Annotations for keyframes are provided at 2 Hz. Each scene has a duration of 20 seconds, resulting in 28,130 and 6,019 frames for the training and validation sets, respectively.

\paragraph{Metric.} We assess the quality of HD semantic learning using two metrics: Mean Intersection over Union (mIoU) and Mean Average Precision (mAP), as presented in HDMapNet ~\cite{li2021hdmapnet}.  In accordance with the methodology detailed in HDMapNet, we evaluate three static map elements:road boundary, lane divider, and pedestrian crossing on the nuScenes dataset.

\subsection{Implementation Details}
\paragraph{Base models.} We primarily perform our experiments using the BEVFormer model\cite{li2022bevformer} (the version excluding the temporal aspect), selected for its strength in BEV feature extraction abilities and its exceptional performance in map semantic segmentation. To validate the broad applicability of our methods, as shown in Table \ref{tab:basemodel_IoU} and \ref{tab:vectormapnet}, we incorporate our NMP paradigm into four recently proposed camera-based map semantic segmentation and detection methods, which serve as our baseline models: HDMapNet\cite{li2021hdmapnet}, LSS\cite{philion2020lift}, BEVFormer~\cite{li2022bevformer}, and VectorMapNet~\cite{liu2022vectormapnet}. Each of these methods implements distinct 2D–3D feature-lifting strategies: MLP-based unprojection is adopted by HDMapNet, depth-based unprojection by LSS, geometry-aware transformer-like models by BEVFormer, and homography-based unprojection by VectorMapNet. For the comparisons presented in Table ~\ref{tab:longshortterm} and Table ~\ref{tab:grid_size}, we only use the GRU fusion module.

\paragraph{C2P Attention.} For all linear layers within the current-to-prior attention module, we set the dimension of the features to 256. For patching, we use a patch size of 10 $\times$ 10, corresponding to a 3m $\times$ 3m area in BEV. This setting preserves local spatial information while conserving parameters.

\paragraph{Global Map Resolution.}
\label{subsec:hyper_param}
We use a default map resolution of 0.3m for the rasterized neural map priors for all experiments and conduct an ablation study on the resolution in Table~\ref{tab:grid_size}. 

\input{table/basemodel_w_map_prior_miou}
\input{table/basemodel_w_map_prior_ap}
\input{table/BEV_range}

\input{table/w_wo_trips}
\input{table/weather}
\subsection{Neural Map Prior Helps Online Map Inference}
\begin{figure*}[!t]
\centering
\includegraphics[width=1.0\linewidth]{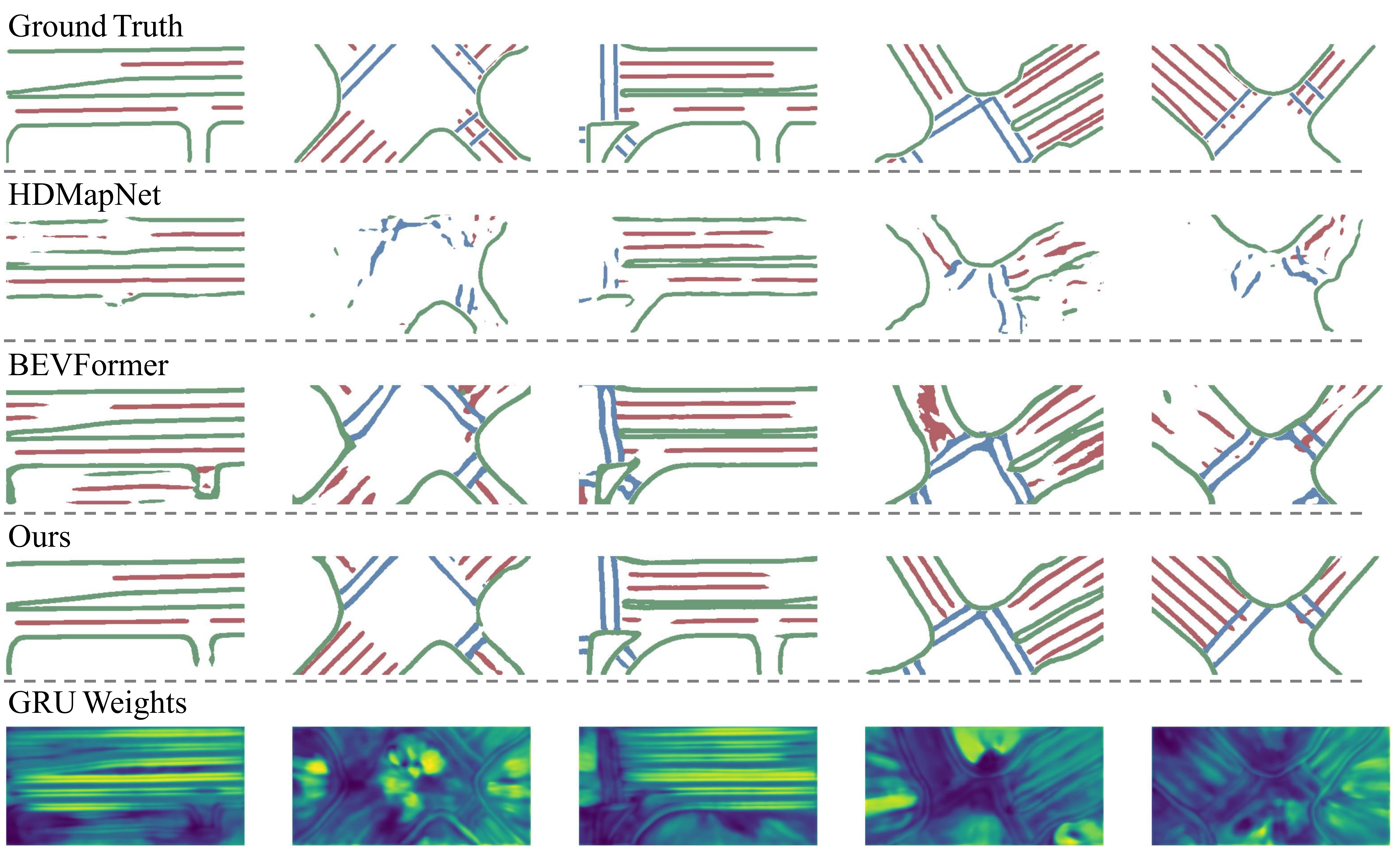}
\caption{\textbf{Qualitative results.} From the first to the fifth row: Ground truth, HDMapNet, BEVFormer, BEVFormer with Neural Map Prior and GRU weights. We also visualize ${z_t}$, the attention map of the last step of the GRU fusion process. The model learns to selectively combine current and prior map features: specifically, when the prediction quality of the current frame is good, the network tends to learn a larger $z_t$, assigning more weight to the current feature; when the prediction quality of the current frame is poor, usually at intersections or locations farther away from the ego-vehicle, the network tends to learn a smaller $z_t$ for the prior feature.}
  \label{fig:qualitative}
\end{figure*}

In this section, we show that the effectiveness of NMP is agnostic to various model architectures and evaluation metrics. To illustrate this, we integrate NMP into the aforementioned four base models: HDMapNet, LSS, BEVFormer, and VectorMapNet. We use the same hyperparameter settings as in their original designs. During training, we freeze all the modules before the BEV features and only train the C2P attention module, the GRU, the local PE, and the decoder. For testing, all samples are arranged chronologically. As evidenced in Table~\ref{tab:basemodel_IoU} and Table~\ref{tab:vectormapnet}, NMP consistently improves map segmentation and detection performance compared to the baseline models. Qualitative results are shown in Figure~\ref{fig:qualitative}. These findings suggest that NMP is a generic approach that can potentially be applied to other mapping frameworks.

\subsection{Neural Map Prior Helps to See Further}
Conventional maps used in autonomous driving systems provide crucial information about roads extending beyond the line of sight, aiding in navigation, planning, and informed decision-making. However, the recent adoption of onboard cameras for online map prediction as an alternative approach has introduced a limitation in the prediction range. This limitation arises due to the low resolution of distant areas in the captured images. To overcome this limitation, our proposed NMP enables an extended reach for online map prediction. Specifically, the NMP leverages prior history information generated by other trips, encapsulating rich contextual details about the scenes and significantly augmenting the capabilities of online map prediction. This enhancement is demonstrated in Table~\ref{tab:bevrange}, which consistently shows improved segmentation results compared to the baseline methods across various BEV ranges, including $60 m \times 30 m$, $100 m \times 100 m$, and $160 m \times 100 m$.

\input{table/method_new.tex}

\subsection{Inter-trip Fusion is better than Intra-trip Fusion}
In Table~\ref{tab:longshortterm}, we show the effectiveness of intra-trip fusion versus inter-trip fusion for map construction. Intra-trip refers to the scenario where the fusion is limited to a single traversal, while the inter-trip fusion model uses map priors generated from multiple traversals at the same location. The findings indicate that the integration of multi-traversal prior information is more helpful for accurate map construction, highlighting the significance of using multiple traversals.

\subsection{Neural Map Prior is more helpful under Adverse Weather Conditions}
Autonomous vehicles face challenges when driving in bad weather conditions or low light conditions, such as rain or night driving, which may impede accurate road information identification. However, our method, NMP, captures and retains the road's appearance under optimal weather and lighting conditions, thereby equipping the vehicle with enhanced and reliable information for precise road perception during current trips. As shown in Table~\ref{tab:weather}, the application of NMP in challenging conditions, including rain and night-time driving, leads to more substantial improvements compared to normal weather scenarios. This indicates that our perception model effectively leverages the necessary information from the NMP to handle bad weather situations. However, given the smaller sample size and the limited prior trip data available for this sample, the improvements were less significant under night-rain conditions.

\subsection{Ablation Studies on Fusion Components}
\label{subsec:ablation}
\input{table/global_grid_size}

\input{table/new_split}

\paragraph{GRU, C2P Attention and Local Position Embedding.}
In this section, we evaluate the effectiveness of the components proposed in Section~\ref{sec:method}. For the sake of comparison, we introduce a simple fusion baseline, termed Moving Average (MA). In this context, the C2P Attention and GRU are replaced with a moving average fusion function. The corresponding update rule can be represented as follows:
\begin{equation}
p_t^l = \alpha o + (1-\alpha)o_{t-1}^l,
\end{equation}
where $\alpha$ denotes a manually searched ratio, and other notations are defined in Section~\ref{subsec:GRU}.
Although both GRU and MA showcase comparable performance enhancements as updating modules, GRU is preferred due to the elimination of manual parameter searches required in MA.
Both GRU and CA act as effective feature fusion modules, resulting in substantial performance enhancements. The slight edge of C2P attention over GRU indicates that the transformer architecture holds a minor advantage in fusing prior feature contexts.
Comparing C to E and F to G in Table~\ref{tab:fusioncomp}, we observe that local PE increases the IoU of the crossing by 2.67 and 2.72, respectively. This suggests that local PE has improved feature fusion, particularly in the challenging category of pedestrian crossings. Local PE enables the model to extract additional information from the map prior, thereby complementing current observations. In comparisons of C to F and E to G in Table~\ref{tab:fusioncomp}, C2P Attention increases the IoU of the lane divider by 1.83 and 2.05, respectively, highlighting its effectiveness in handling lane structures. The attention mechanism extracts relevant features based on the spatial context, leading to a more accurate understanding of dividers and boundary structures. Overall, the ablation study confirms the effectiveness of all three proposed components for feature fusion and updating.

\paragraph{Map Resolution.}
We investigate the impact of different resolutions of global neural priors on the effectiveness of online map learning in Table~\ref{tab:grid_size}. High resolutions are preferred to preserve details on the map. However, there is a trade-off between storage and resolution. Our experiments achieved good performance with an appropriate resolution of 0.3m.

\subsection{Dataset Re-split}
In the original split of the nuScenes dataset, some samples lack historical traversals. We adopt an approach similar to the one presented in Hindsight~\cite{you2022hindsight}, to re-split the trips in Boston, and name it as \textit{Boston split}. The Boston split ensures that each sample includes at least one historical trip, while the training and test samples are geographically disjoint.
To estimate the proximity of two samples, we calculate the areal overlap, specifically IoU in the bird's-eye view, between the field of view of the two traversals.
This approach results in 7354 training samples and 6504 test samples. The comparison of model performance on the original split and Boston split is shown in Table \ref{tab:split}. The improvement of NMP observed on the Boston split is greater compared to the original split.


\subsection{Map Tiles}
\label{subsec:exp_tile}
We use map tile as the storage format for our global neural map prior. In urban environments, buildings generally occupy a substantial portion of the area, whereas road-related regions account for a smaller part. To prevent the map's storage size from expanding excessively in proportion to the physical scale of the city, we design a storage structure that divides the city into sparse pieces indexed by their physical coordinates. It consumes 70\% less memory space than dense tiles.
Furthermore, each vehicle does not need to store the entire city map; instead, it can download map tiles on demand.
The trained model remains fixed, but these map tiles are updated, integrated, and uploaded to the cloud asynchronously. As more trip data is collected over time, the map prior becomes broader and of better quality.

%% file: table/basemodel_w_map_prior_miou.tex
\begin{table}[t]
    \small\centering
    \caption{\textbf{Quantitative analysis of map segmentation.} The performance of online map segmentation methods and their NMP versions on the nuScenes validation set. By adding prior knowledge, NMP consistently improves these methods. (* HDMapNet remains the same as in the original work, while LSS uses the same backbone as BEVFormer.)}
    \resizebox{1.0\linewidth}{!}{
    \begin{tabular}{lcccc}
            \Xhline{3\arrayrulewidth}
    		\multirow{2}{*}{Model}&\multicolumn{4}{c}{mIoU}\\
            \cline{2-5}
    		&Divider&Crossing&Boundary&All\\
    		\midrule
    		HDMapNet&41.04&16.23&40.93&32.73\\
            HDMapNet + NMP&\textbf{44.15}&\textbf{20.95}&\textbf{46.07}&\textbf{37.05}\\
            \cdashline{1-5}
            \rowcolor{lightgreen}
            $\triangle$ mIoU&\textcolor{gray}{+3.11}&\textcolor{gray}{+4.72}&\textcolor{gray}{+5.14}&\textcolor{gray}{+4.32}\\
            \midrule
            LSS$^*$&45.19&26.90&47.27&39.78\\ 
            LSS$^*$ + NMP&\textbf{50.20}&\textbf{30.66}&\textbf{53.56}&\textbf{44.80}\\
            \cdashline{1-5}
            \rowcolor{lightgreen}
            $\triangle$ mIoU&\textcolor{gray}{+5.01}&\textcolor{gray}{+3.76}&\textcolor{gray}{+6.29}&\textcolor{gray}{+5.02}\\
            \midrule
		BEVFormer$^*$&49.51&28.85&50.67&43.01\\
		BEVFormer$^*$ + NMP&\textbf{55.01}&\textbf{34.09}&\textbf{56.52}&\textbf{48.54}\\
            \cdashline{1-5}
            \rowcolor{lightgreen}
            $\triangle$ mIoU&\textcolor{gray}{+5.50}&\textcolor{gray}{+5.24}&\textcolor{gray}{+5.95}&\textcolor{gray}{+5.53}\\
            \Xhline{3\arrayrulewidth}
    \end{tabular}
    }
    \vspace{-0.2cm}
    \label{tab:basemodel_IoU}
\end{table}

%% file: table/basemodel_w_map_prior_ap.tex
\begin{table}[!t]
    \small\centering
    \caption{\textbf{Quantitative analysis of vectorized map detection.} By adding prior knowledge, the NMP boosts the performance of VectorMapNet.}
    \resizebox{1.0\linewidth}{!}{
    \begin{tabular}{lcccc}
            \Xhline{3\arrayrulewidth}
		\multirow{2}{*}{Model}&\multicolumn{4}{c}{Average Precision}\\
            \cline{2-5}
		&$AP_{Divider}$& $AP_{Crossing}$&$AP_{Boundary}$&mAP\\
		\midrule
            VectorMapNet&47.3&36.1&39.3&40.9\\
            VectorMapNet + NMP&\textbf{49.6}&\textbf{42.9}&\textbf{41.9}&\textbf{44.8}\\
            \cdashline{1-5}
            \rowcolor{lightgreen}
            $\triangle$ AP&\textcolor{gray}{+2.3}&\textcolor{gray}{+6.8}&\textcolor{gray}{+2.6}&\textcolor{gray}{+3.9}\\
            \Xhline{3\arrayrulewidth}
    \end{tabular}
    }
    \label{tab:vectormapnet}
        \vspace{-0.2cm}
\end{table}


%% file: table/bev_range.tex
\begin{table}[!t]
    \small\centering
        \caption{\textbf{Comparison of model performance at different BEV ranges.} As the perception range increases, the online method performance declines.; NMP significantly improves the results.}
	\resizebox{1.0\linewidth}{!}{
		\begin{tabular}{lccccc}
            \Xhline{3\arrayrulewidth}
		\multirow{2}{*}{BEV Range}&\multirow{2}{*}{+ NMP}&\multicolumn{4}{c}{mIoU}\\
            \cline{3-6}
		&&Divider& Crossing&Boundary&All\\
		\midrule
		\multirow{2}{*}{$60m \times\ 30m$}\ &\xmark&49.51&28.85&50.67&43.01\\
		&\checkmark&\textbf{55.01}&\textbf{34.09}&\textbf{56.52}&\textbf{48.54}\\
            \cdashline{1-6}
            $\triangle$ mIoU&&\textcolor{gray}{+5.50}&\textcolor{gray}{+5.24}&\textcolor{gray}{+5.95}&\textcolor{gray}{+5.53}\\
		\midrule
		\multirow{2}{*}{$100m \times\ 100m$}\ &\xmark&43.41&29.07&56.57&43.01\\
		&\checkmark&\textbf{49.51}&\textbf{32.67}&\textbf{59.94}&\textbf{47.37}\\
            \hdashline
            $\triangle$ mIoU&&\textcolor{gray}{+6.10}&\textcolor{gray}{+3.60}&\textcolor{gray}{+3.60}&\textcolor{gray}{+4.36}\\
		\midrule
		\multirow{2}{*}{$160m \times\ 100m$}\ &\xmark&41.21&26.42&51.74&39.79\\ 
		&\checkmark&\textbf{46.85}&\textbf{29.25}&\textbf{57.22}&\textbf{44.44}\\
          \hdashline
            $\triangle$ mIoU&&\textcolor{gray}{+5.64}&\textcolor{gray}{+2.83}&\textcolor{gray}{+5.48}&\textcolor{gray}{+4.65}\\
            \Xhline{3\arrayrulewidth}
		\end{tabular}
	}
\label{tab:bevrange}
\end{table}

%% file: table/w_wo_trips.tex
\begin{table}[!t]
\setlength{\tabcolsep}{4.5pt}
\small\centering
\caption{\textbf{Comparison of intra-trip fusion and inter-trip fusion.}}
	\resizebox{1.0\linewidth}{!}{
		\begin{tabular}{lcccc}
            \Xhline{3\arrayrulewidth}
		\multirow{2}{*}{Intra or Inter Trips}&\multicolumn{4}{c}{mIoU}\\
            \cline{2-5}
		& Divider& Crossing&Boundary&All\\
		\midrule
		Baseline&49.51&28.85&50.67&43.01\\
		Intra-trip fusion &51.87&30.34&53.74&45.31(+2.30)\\
            \rowcolor{LightCyan}
            Inter-trip fusion&\textbf{53.41}&\textbf{31.92}&\textbf{55.15}&\textbf{46.82(+3.81)}\\
            \Xhline{3\arrayrulewidth}
		\end{tabular}
	}
\label{tab:longshortterm}
\end{table}

%% file: table/weather.tex
\begin{table}[!h]
    \small\centering
    \setlength{\tabcolsep}{4.5pt}
        \caption{\textbf{Performance in adverse weather conditions.}}
          \resizebox{1.0\linewidth}{!}{
          \begin{tabular}{lccccc}
            \Xhline{3\arrayrulewidth}
		\multirow{2}{*}{Weather}&\multirow{2}{*}{+ NMP}&\multicolumn{4}{c}{mIoU}\\
            \cline{3-6}
		&&Divider& Crossing&Boundary&All\\
		\midrule
		\multirow{2}{*}{Rain}&\xmark&50.25&26.90&44.54&40.56\\
		&\checkmark&\textbf{54.64}&\textbf{30.62}&\textbf{54.19}&\textbf{46.48}\\
            \cdashline{1-6}
		$\triangle$ mIoU&&\textcolor{gray}{+4.39}&\textcolor{gray}{+3.72}&\textcolor{gray}{+9.65}&\textcolor{gray}{+5.92}\\
		\midrule
		\multirow{2}{*}{Night}&\xmark&51.02&21.17&48.99&40.39\\
		&\checkmark&\textbf{ 54.66}&\textbf{33.78}&\textbf{55.92}&\textbf{48.12}\\
            \cdashline{1-6}
		$\triangle$ mIoU&&\textcolor{gray}{+3.64}&\textcolor{gray}{+12.61}&\textcolor{gray}{+6.93}&\textcolor{gray}{+7.73}\\
		\midrule
		\multirow{2}{*}{NightRain}&\xmark&55.76&00.00&47.60&34.45\\
		&\checkmark&\textbf{61.22}&00.00&\textbf{50.84}&\textbf{37.35}\\
            \cdashline{1-6}
		$\triangle$ mIoU&&\textcolor{gray}{+5.46}&\textcolor{gray}{+00.00}&\textcolor{gray}{+3.24}&\textcolor{gray}{+2.90}\\
		\midrule
		\multirow{2}{*}{Normal}&\xmark&49.27&29.49&52.11&43.62\\
		&\checkmark&\textbf{53.46}&\textbf{35.27}&\textbf{57.75}&\textbf{48.82}\\
            \cdashline{1-6}
		$\triangle$ mIoU&&\textcolor{gray}{+4.19}&\textcolor{gray}{+5.78}&\textcolor{gray}{+5.64}&\textcolor{gray}{+5.20}\\
            \Xhline{3\arrayrulewidth}
		\end{tabular}
	}
\label{tab:weather}
\end{table}


%% file: table/method_new.tex
\begin{table*}[!t]
    \small\centering
    \caption{\textbf{Ablation on the fusion components.} MA stands for Moving Average. Local PE stands for the positional embedding proposed in \S~\ref{subsec:localinference}. CA stands for the C2P Attention proposed in \S~\ref{subsec:localinference} and GRU stands for gated recurrent units proposed in \S~\ref{subsec:GRU}. 
    }
		\begin{tabular}{ccccccccc}
                \Xhline{3\arrayrulewidth}
    		& \multicolumn{4}{c}{Component} &\multicolumn{4}{c}{mIoU}\\
                \cline{1-5} \cline{6-9}
    		Name&MA&GRU&Local PE&CA&Divider& Crossing&Boundary&All\\
    		\midrule
                A&&&&&49.51&28.85&50.67&43.01\\
                B&\checkmark&&&&52.19(+2.68)&33.70(+4.85)&55.34(+4.67)&47.07(+4.06)\\
    		C&&\checkmark &&&53.22(+3.71)&31.46(+2.61)  &55.93(+5.26)&46.87(+3.86)\\
    		D&&&&\checkmark&53.25(+3.74)&33.13(+4.28)&55.15(+4.48)&47.17(+4.16)\\
                E&&\checkmark&\checkmark&&52.96(+3.45)&\textbf{34.13(+5.28)}&56.14(+5.47)&47.74(+4.73)\\
    		F&&\checkmark&&\checkmark&\textbf{55.05(+3.74)}&31.37(+2.52)&56.19(+5.52)&47.53(+4.52)\\
    		G&&\checkmark&\checkmark&\checkmark& 55.01(+5.50)&34.09(+5.24)&\textbf{56.52(+5.85)} & \textbf{48.54(+5.53)}\\
                \Xhline{3\arrayrulewidth}
           \end{tabular}
\label{tab:fusioncomp}
\end{table*}

%% file: table/global_grid_size.tex
\begin{table}[!t]
    \small\centering
    \setlength{\tabcolsep}{4.5pt}
    \caption{\textbf{Ablation on the global map resolution.} 0.3m $\times$ 0.3m is a good design choice that balances storage size and accuracy.}
	\resizebox{1.0\linewidth}{!}{
		\begin{tabular}{lcccc}
            \Xhline{3\arrayrulewidth}
		\multirow{2}{*}{NMP Grid Resolution}&\multicolumn{4}{c}{mIoU}\\
            \cline{2-5}
		&Divider & Crossing &Boundary &All\\
		\midrule
		Baseline &49.51 &28.85 &50.67 &43.01\\
            \rowcolor{LightCyan}
		0.3m $\times$ 0.3m&\textbf{53.22}&31.46&\textbf{55.93}&\textbf{46.87(+3.86)}\\
            0.6m $\times$ 0.6m&52.42&\textbf{31.63}&54.74&46.26(+3.25)\\
            \rowcolor[gray]{0.95}
            1.2m $\times$ 1.2m&51.36&30.24&52.78&44.79(+1.78)\\
            \Xhline{3\arrayrulewidth}
		\end{tabular}
	}
\label{tab:grid_size}
\end{table}

%% file: table/new_split.tex
\begin{table}[!t]
    \small\centering
        \caption{\textbf{Performance on Boston split.} The original split contains unbalanced historical trips for the training and validation sets; Boston split is more balanced.}
	\resizebox{1.0\linewidth}{!}{
		\begin{tabular}{lccccc}
            \Xhline{3\arrayrulewidth}
		\multirow{2}{*}{Data Split}&\multirow{2}{*}{+ NMP}&\multicolumn{4}{c}{mIoU}\\
            \cline{3-6}
		&&Divider& Crossing&Boundary&All\\
		\midrule
		\multirow{2}{*}{Boston Split}&\xmark&26.35&15.32&25.06&22.24\\
&\checkmark&\textbf{33.04}&\textbf{21.72}&\textbf{32.63}&\textbf{29.13}\\
            \cdashline{1-6}
            $\triangle$ mIoU&&\textcolor{gray}{+6.69}&\textcolor{gray}{+6.40}&\textcolor{gray}{+7.57}&\textcolor{gray}{+6.89}\\
            \midrule
            \multirow{2}{*}{Original Split}&\xmark&49.51&28.85&50.67&43.01\\
            &\checkmark&\textbf{55.01}&\textbf{34.09}&\textbf{56.52}&\textbf{48.54}\\
            \cdashline{1-6}
            $\triangle$ mIoU&&\textcolor{gray}{+5.50}&\textcolor{gray}{+5.24}&\textcolor{gray}{+5.95}&\textcolor{gray}{+5.53}\\
            \Xhline{3\arrayrulewidth}
		\end{tabular}
	}
 
\label{tab:split}
\end{table}


%% file: conclusion.tex
\section{Conclusion}
In this paper, we introduce a novel system, Neural Map Prior, which is designed to enhance online learning of HD semantic maps. NMP involves the joint execution of global map prior updates and local map inference for each frame in an incremental manner. A comprehensive analysis on the nuScenes dataset demonstrates that NMP improves online map inference performance, especially in challenging weather conditions and extended prediction horizons.
Future work includes learning more semantic map elements and 3D maps.

%% file: main.bbl
\begin{thebibliography}{10}\itemsep=-1pt

\bibitem{besl1992method}
Paul~J Besl and Neil~D McKay.
\newblock Method for registration of 3-d shapes.
\newblock In {\em Sensor fusion IV: control paradigms and data structures},
  volume 1611, pages 586--606. Spie, 1992.

\bibitem{biber2003normal}
Peter Biber and Wolfgang Stra{\ss}er.
\newblock The normal distributions transform: A new approach to laser scan
  matching.
\newblock In {\em Proceedings 2003 IEEE/RSJ International Conference on
  Intelligent Robots and Systems (IROS 2003)(Cat. No. 03CH37453)}, volume~3,
  pages 2743--2748. IEEE, 2003.

\bibitem{caesar2020nuscenes}
Holger Caesar, Varun Bankiti, Alex~H Lang, Sourabh Vora, Venice~Erin Liong,
  Qiang Xu, Anush Krishnan, Yu Pan, Giancarlo Baldan, and Oscar Beijbom.
\newblock nuscenes: A multimodal dataset for autonomous driving.
\newblock In {\em Proceedings of the IEEE/CVF conference on computer vision and
  pattern recognition}, pages 11621--11631, 2020.

\bibitem{can2020understanding}
Yigit~Baran Can, Alexander Liniger, Ozan Unal, Danda Paudel, and Luc Van~Gool.
\newblock Understanding bird's-eye view semantic hd-maps using an onboard
  monocular camera.
\newblock {\em arXiv preprint arXiv:2012.03040}, 2020.

\bibitem{chen2022futr3d}
Xuanyao Chen, Tianyuan Zhang, Yue Wang, Yilun Wang, and Hang Zhao.
\newblock Futr3d: A unified sensor fusion framework for 3d detection.
\newblock {\em arXiv preprint arXiv:2203.10642}, 2022.

\bibitem{chung2014empirical}
Junyoung Chung, Caglar Gulcehre, KyungHyun Cho, and Yoshua Bengio.
\newblock Empirical evaluation of gated recurrent neural networks on sequence
  modeling.
\newblock {\em arXiv preprint arXiv:1412.3555}, 2014.

\bibitem{dellaert2012factor}
Frank Dellaert.
\newblock Factor graphs and gtsam: A hands-on introduction.
\newblock Technical report, Georgia Institute of Technology, 2012.

\bibitem{jiangLocalImplicitGrid2020}
Chiyu Jiang, Avneesh Sud, Ameesh Makadia, Jingwei Huang, Matthias Nie{\ss}ner,
  and Thomas Funkhouser.
\newblock Local implicit grid representations for 3d scenes.
\newblock In {\em CVPR}, 2020.

\bibitem{jiao2018machine}
Jialin Jiao.
\newblock Machine learning assisted high-definition map creation.
\newblock In {\em 2018 IEEE 42nd Annual Computer Software and Applications
  Conference (COMPSAC)}, volume~1, pages 367--373. IEEE, 2018.

\bibitem{lawson1995solving}
Charles~L Lawson and Richard~J Hanson.
\newblock {\em Solving least squares problems}.
\newblock SIAM, 1995.

\bibitem{li2021hdmapnet}
Qi Li, Yue Wang, Yilun Wang, and Hang Zhao.
\newblock Hdmapnet: A local semantic map learning and evaluation framework.
\newblock {\em arXiv preprint arXiv:2107.06307}, 2021.

\bibitem{li2022bevformer}
Zhiqi Li, Wenhai Wang, Hongyang Li, Enze Xie, Chonghao Sima, Tong Lu, Qiao Yu,
  and Jifeng Dai.
\newblock Bevformer: Learning bird's-eye-view representation from multi-camera
  images via spatiotemporal transformers.
\newblock {\em arXiv preprint arXiv:2203.17270}, 2022.

\bibitem{liao2022maptr}
Bencheng Liao, Shaoyu Chen, Xinggang Wang, Tianheng Cheng, Qian Zhang, Wenyu
  Liu, and Chang Huang.
\newblock Maptr: Structured modeling and learning for online vectorized hd map
  construction.
\newblock {\em arXiv preprint arXiv:2208.14437}, 2022.

\bibitem{liu2020neural}
Lingjie Liu, Jiatao Gu, Kyaw~Zaw Lin, Tat-Seng Chua, and Christian Theobalt.
\newblock Neural sparse voxel fields.
\newblock In {\em NeurIPS}, 2020.

\bibitem{liu2022vectormapnet}
Yicheng Liu, Yue Wang, Yilun Wang, and Hang Zhao.
\newblock Vectormapnet: End-to-end vectorized hd map learning.
\newblock {\em arXiv preprint arXiv:2206.08920}, 2022.

\bibitem{liu2021swin}
Ze Liu, Yutong Lin, Yue Cao, Han Hu, Yixuan Wei, Zheng Zhang, Stephen Lin, and
  Baining Guo.
\newblock Swin transformer: Hierarchical vision transformer using shifted
  windows.
\newblock In {\em Proceedings of the IEEE/CVF International Conference on
  Computer Vision}, pages 10012--10022, 2021.

\bibitem{lu2019monocular}
Chenyang Lu, Marinus Jacobus~Gerardus van~de Molengraft, and Gijs Dubbelman.
\newblock Monocular semantic occupancy grid mapping with convolutional
  variational encoder--decoder networks.
\newblock {\em IEEE Robotics and Automation Letters}, 4(2):445--452, 2019.

\bibitem{mattyus2015enhancing}
Gellert Mattyus, Shenlong Wang, Sanja Fidler, and Raquel Urtasun.
\newblock Enhancing road maps by parsing aerial images around the world.
\newblock In {\em Proceedings of the IEEE international conference on computer
  vision}, pages 1689--1697, 2015.

\bibitem{mattyus2016hd}
Gell{\'e}rt M{\'a}ttyus, Shenlong Wang, Sanja Fidler, and Raquel Urtasun.
\newblock Hd maps: Fine-grained road segmentation by parsing ground and aerial
  images.
\newblock In {\em Proceedings of the IEEE Conference on Computer Vision and
  Pattern Recognition}, pages 3611--3619, 2016.

\bibitem{meschederOccupancyNetworksLearning2018}
Lars Mescheder, Michael Oechsle, Michael Niemeyer, Sebastian Nowozin, and
  Andreas Geiger.
\newblock Occupancy networks: Learning 3d reconstruction in function space.
\newblock In {\em CVPR}, 2019.

\bibitem{pan2020cross}
Bowen Pan, Jiankai Sun, Ho~Yin~Tiga Leung, Alex Andonian, and Bolei Zhou.
\newblock Cross-view semantic segmentation for sensing surroundings.
\newblock {\em IEEE Robotics and Automation Letters}, 5(3):4867--4873, 2020.

\bibitem{parkDeepSDFLearningContinuous2019}
Jeong~Joon Park, Peter Florence, Julian Straub, Richard Newcombe, and Steven
  Lovegrove.
\newblock {{DeepSDF}}: {{Learning Continuous Signed Distance Functions}} for
  {{Shape Representation}}.
\newblock In {\em CVPR}, 2019.

\bibitem{philion2020lift}
Jonah Philion and Sanja Fidler.
\newblock Lift, splat, shoot: Encoding images from arbitrary camera rigs by
  implicitly unprojecting to 3d.
\newblock In {\em European Conference on Computer Vision}, pages 194--210.
  Springer, 2020.

\bibitem{pomerleau2015review}
Fran{\c{c}}ois Pomerleau, Francis Colas, Roland Siegwart, et~al.
\newblock A review of point cloud registration algorithms for mobile robotics.
\newblock {\em Foundations and Trends{\textregistered} in Robotics},
  4(1):1--104, 2015.

\bibitem{roddick2020predicting}
Thomas Roddick and Roberto Cipolla.
\newblock Predicting semantic map representations from images using pyramid
  occupancy networks.
\newblock In {\em Proceedings of the IEEE/CVF Conference on Computer Vision and
  Pattern Recognition}, pages 11138--11147, 2020.

\bibitem{rong2020lgsvl}
Guodong Rong, Byung~Hyun Shin, Hadi Tabatabaee, Qiang Lu, Steve Lemke,
  M{\=a}rti{\c{n}}{\v{s}} Mo{\v{z}}eiko, Eric Boise, Geehoon Uhm, Mark Gerow,
  Shalin Mehta, et~al.
\newblock Lgsvl simulator: A high fidelity simulator for autonomous driving.
\newblock {\em arXiv preprint arXiv:2005.03778}, 2020.

\bibitem{saha2022translating}
Avishkar Saha, Oscar Mendez, Chris Russell, and Richard Bowden.
\newblock Translating images into maps.
\newblock In {\em 2022 International Conference on Robotics and Automation
  (ICRA)}, pages 9200--9206. IEEE, 2022.

\bibitem{saitoPIFuPixelAlignedImplicit2019}
Shunsuke Saito, Zeng Huang, Ryota Natsume, Shigeo Morishima, Angjoo Kanazawa,
  and Hao Li.
\newblock {{PIFu}}: {{Pixel}}-{{Aligned Implicit Function}} for
  {{High}}-{{Resolution Clothed Human Digitization}}.
\newblock In {\em ICCV}, 2019.

\bibitem{segal2009Generalized}
Aleksandr Segal, Dirk Haehnel, and Sebastian Thrun.
\newblock Generalized-icp.
\newblock In {\em Robotics: science and systems}, volume~2, page 435. Seattle,
  WA, 2009.

\bibitem{sun2021neuralrecon}
Jiaming Sun, Yiming Xie, Linghao Chen, Xiaowei Zhou, and Hujun Bao.
\newblock Neuralrecon: Real-time coherent 3d reconstruction from monocular
  video.
\newblock In {\em Proceedings of the IEEE/CVF Conference on Computer Vision and
  Pattern Recognition}, pages 15598--15607, 2021.

\bibitem{wang2016torontocity}
Shenlong Wang, Min Bai, Gellert Mattyus, Hang Chu, Wenjie Luo, Bin Yang, Justin
  Liang, Joel Cheverie, Sanja Fidler, and Raquel Urtasun.
\newblock Torontocity: Seeing the world with a million eyes.
\newblock {\em arXiv preprint arXiv:1612.00423}, 2016.

\bibitem{wang2015holistic}
Shenlong Wang, Sanja Fidler, and Raquel Urtasun.
\newblock Holistic 3d scene understanding from a single geo-tagged image.
\newblock In {\em Proceedings of the IEEE Conference on Computer Vision and
  Pattern Recognition}, pages 3964--3972, 2015.

\bibitem{wang2022detr3d}
Yue Wang, Vitor~Campagnolo Guizilini, Tianyuan Zhang, Yilun Wang, Hang Zhao,
  and Justin Solomon.
\newblock Detr3d: 3d object detection from multi-view images via 3d-to-2d
  queries.
\newblock In {\em Conference on Robot Learning}, pages 180--191. PMLR, 2022.

\bibitem{yang2018hdnet}
Bin Yang, Ming Liang, and Raquel Urtasun.
\newblock Hdnet: Exploiting hd maps for 3d object detection.
\newblock In {\em Conference on Robot Learning}, pages 146--155. PMLR, 2018.

\bibitem{yang2018robust}
Sheng Yang, Xiaoling Zhu, Xing Nian, Lu Feng, Xiaozhi Qu, and Teng Ma.
\newblock A robust pose graph approach for city scale lidar mapping.
\newblock In {\em 2018 IEEE/RSJ International Conference on Intelligent Robots
  and Systems (IROS)}, pages 1175--1182. IEEE, 2018.

\bibitem{yang2021projecting}
Weixiang Yang, Qi Li, Wenxi Liu, Yuanlong Yu, Yuexin Ma, Shengfeng He, and Jia
  Pan.
\newblock Projecting your view attentively: Monocular road scene layout
  estimation via cross-view transformation.
\newblock In {\em Proceedings of the IEEE/CVF Conference on Computer Vision and
  Pattern Recognition}, pages 15536--15545, 2021.

\bibitem{yarivMultiviewNeuralSurface2020}
Lior Yariv, Yoni Kasten, Dror Moran, Meirav Galun, Matan Atzmon, Basri Ronen,
  and Yaron Lipman.
\newblock Multiview neural surface reconstruction by disentangling geometry and
  appearance.
\newblock In {\em NeurIPS}, 2020.

\bibitem{you2022hindsight}
Yurong You, Katie~Z Luo, Xiangyu Chen, Junan Chen, Wei-Lun Chao, Wen Sun,
  Bharath Hariharan, Mark Campbell, and Kilian~Q Weinberger.
\newblock Hindsight is 20/20: Leveraging past traversals to aid 3d perception.
\newblock {\em arXiv preprint arXiv:2203.11405}, 2022.

\bibitem{yu2015semantic}
Fisher Yu, Jianxiong Xiao, and Thomas Funkhouser.
\newblock Semantic alignment of lidar data at city scale.
\newblock In {\em Proceedings of the IEEE Conference on Computer Vision and
  Pattern Recognition}, pages 1722--1731, 2015.

\bibitem{zhou2022cross}
Brady Zhou and Philipp Kr{\"a}henb{\"u}hl.
\newblock Cross-view transformers for real-time map-view semantic segmentation.
\newblock {\em arXiv preprint arXiv:2205.02833}, 2022.

\end{thebibliography}
